\documentclass[10pt]{article}
\usepackage{environ}
\usepackage[verbose=true,letterpaper,margin=0.6in]{geometry}

\usepackage[T1]{fontenc}
\usepackage{newtxtext}
\usepackage{newtxmath}
\usepackage{bm}
\usepackage{amsmath}
\usepackage{mathtools}
\usepackage{graphicx}
\usepackage{algorithm}
\usepackage{algpseudocode}
\usepackage{booktabs}
\usepackage{multirow}
\usepackage[font=small,labelfont=bf]{caption}
\usepackage{authblk}
\usepackage{natbib}
\bibpunct[, ]{(}{)}{,}{a}{}{,}%
\usepackage[colorlinks=true,linkcolor=blue,citecolor=blue,urlcolor=blue]{hyperref}

\newcommand\m[1]{\ensuremath{\mathcal{#1}}}
\newcommand{\feature}{\bm{x}}
\newcommand{\cost}{\bm{c}}
\newcommand{\predcost}{\bm{\hat{c}}}
\newcommand{\decisionvar}{\bm{w}}

\newcommand{\feasspace}{\m{W}}
\newcommand{\spoplus}{\ell_{\mathrm{SPO}+}}
\newcommand{\R}{\mathbb{R}}
\newcommand{\E}{\mathbb{E}}
\providecommand{\argmin}{\operatorname*{arg\,min}}

\DeclareMathOperator{\CVaR}{CVaR}
\newcommand{\restro}[1]{\mathit{RESTRO}_{\text{#1}}}

\newcommand{\INFORMSnotes}[1]{%
  \if\relax\detokenize{#1}\relax\else\par\smallskip{\footnotesize #1\par}\fi}
\newcommand{\TABLE}[3]{\centering\caption{#1}#2\INFORMSnotes{#3}}

\newcommand{\Halmos}{\mbox{\quad$\square$}}
\let\endnote\footnote
\newtheorem{proposition}{Proposition}
\newtheorem{assumption}{Assumption}
\newtheorem{remark}{Remark}
\newtheorem{lemma}{Lemma} 
\newenvironment{proof}[1]{\par\medskip\noindent\textit{#1}\enskip\ignorespaces}{\par\medskip}

\title{Risk-Averse Decision-Focused Learning:\\ Minimizing CVaR of Regret}

\author[1]{Jayanta Mandi}
\author[2]{Senne Berden}
\author[3]{V\'ictor Bucarey}
\author[4]{Concepci\'on Dom\'inguez}
\author[2]{Tias Guns}

\affil[1]{School of Architecture, Building and Civil Engineering, Loughborough University, UK\authorcr
	{\small\texttt{j.mandi1@lboro.ac.uk}}}
\affil[2]{Department of Computer Science, KU Leuven, Leuven, Belgium\authorcr
	{\small\texttt{\{senne.berden, tias.guns\}@kuleuven.be}}}
\affil[3]{Institute of Engineering Sciences, Universidad de O'Higgins, Rancagua, Chile\authorcr
	{\small\texttt{victor.bucarey@uoh.cl}}}
\affil[4]{Department of Statistics and Operations Research, Universidad de Murcia, Murcia, Spain\authorcr
	{\small\texttt{concepcion.dominguez@um.es}}}

\date{}

\begin{document}
\maketitle

\begin{abstract}
Many combinatorial optimization (CO) problems involve parameters that are uncertain when the problem must be solved. In such settings, the parameters have to be predicted from correlated contextual information before a solution can be computed. \emph{Decision-focused learning} (DFL) addresses this by training the predictive model to produce predictions that optimize downstream decision quality. Most existing DFL methods are risk-neutral and minimize the expected regret. In high-stakes applications this can lead to undesirable outcomes, because decision-makers must also account for the tail risk of decision quality. Prior risk-sensitive approaches typically model uncertainty explicitly and solve a stochastic or robust optimization problem at inference time, which increases computational complexity. We instead propose a DFL framework that achieves risk-sensitive behavior while retaining point predictions and deterministic optimization at inference time. Starting from the dual representation of the Conditional Value-at-Risk (CVaR) and adding entropy regularization, we derive a smooth surrogate for the CVaR of regret. Its gradient induces an instance-dependent reweighting within each minibatch that emphasizes high-risk outcomes, so the approach can turn existing surrogate-loss DFL methods, including the SPO+ loss, into risk-sensitive variants. When instantiated with SPO+, the resulting objective admits a generalization bound on the CVaR of the true regret. Experiments on knapsack, portfolio, shortest-path, and energy-aware scheduling problems show substantial reductions in tail regret when the uncertain parameters exhibit heterogeneous tail behavior.

\end{abstract}


\section{Introduction}\label{sec:intro}

A wide range of real-world decision-making problems can be modeled as combinatorial optimization (CO) problems. Examples include network optimization~\citep{HuangWaterPipe}, production planning~\citep{pochet2006production}, and vehicle routing~\citep{toth2002vehicle}. In practice, these problems often depend on parameters that are unknown at decision time, such as traffic congestion in routing or future costs in production planning. To handle this uncertainty, there has been growing interest in treating such problems as contextual stochastic optimization problems~\citep{contextual2025}. This perspective gives rise to \emph{predict-then-optimize} (PtO) frameworks~\citep{DFLsurvey}, in which a machine learning (ML) model first predicts the unknown parameters from contextual features, and the resulting instantiated optimization problem is then solved.

A straightforward approach to PtO problems is \emph{prediction-focused learning} (PFL), in which the ML model is trained to predict the unknown parameters as accurately as possible with a standard supervised loss such as the mean squared error (MSE). A key limitation of PFL is that it ignores how prediction errors affect the downstream decision, even though not all prediction errors are equally harmful. Consider a routing problem with three possible paths that take 10, 12, and 30 minutes to traverse. Predicting the travel time of the third path accurately is largely irrelevant, as long as it is predicted to be clearly worse than the other two. In contrast, a small error on the first two paths may change the chosen route. Because PFL penalizes errors of equal magnitude equally, it does not align learning with downstream decision quality.

\emph{Decision-focused learning} (DFL)~\citep{DFLsurvey} addresses this limitation by embedding the optimization problem into the training loop of the predictive model. Instead of minimizing the error of the predicted parameters, DFL minimizes a decision loss such as the \emph{regret}, which measures the suboptimality of the decisions induced by the predicted parameters. The model is thus trained specifically to make predictions that lead to good decisions. Gradient-based DFL requires differentiating through the CO problem, which is generally not possible because the optimal solution of a CO problem does not change smoothly with its parameters. A widely used remedy is to train with a surrogate loss~\citep{elmachtoub2022smart, mulamba2020discrete} whose subgradients provide useful descent directions for the regret.

Most existing DFL methods, however, adopt a risk-neutral perspective and minimize the expected regret. In many applications the average regret is not the primary concern. Decision-makers often operate in environments where rare but severe outcomes dominate operational risk. Consider dispatching emergency vehicles over a road network with uncertain travel times: a route that is fastest on average may occasionally incur extreme delays, which can be unacceptable in such a high-stakes setting. In these cases the objective should be to minimize the tail risk rather than the average regret.

Risk-sensitive decision making is commonly modeled with measures such as the Conditional Value-at-Risk (CVaR), which evaluates the average of the worst $(1-\alpha)$ fraction of outcomes. Recent research has started to incorporate such risk measures into PtO. Existing approaches either learn a conditional distribution over the parameters and solve a risk-averse stochastic optimization problem by sample average approximation~\citep{gendfl}, or learn uncertainty or ambiguity sets and solve a robust or distributionally robust optimization problem at inference time~\citep{chenreddy2024end, ma2024differentiable, predictthencalibrate}. Although these approaches target tail risk, they increase computational complexity, because a stochastic or (distributionally) robust optimization problem must be solved every time a decision is made~\citep{chicoisne2018risk}.

Many applications, however, require fast inference, and solving a complex risk-averse optimization problem for every decision may be computationally prohibitive. This motivates the question: \emph{can risk-sensitive decision making be achieved while retaining simple point predictions and deterministic optimization at inference time?}

To answer this question, we start from the dual formulation of CVaR and derive, through entropy regularization, a smooth surrogate for the tail regret. The derivation shows that risk-sensitive DFL can be achieved by a principled, instance-dependent reweighting of subgradients, where the weights follow from the first-order optimality conditions of the regularized objective. Instances with larger risk receive higher weight, while lower-risk instances remain informative rather than being discarded. We call this approach \emph{REweighted Subgradients for Tail Regret Optimization} (RESTRO). RESTRO operates on top of existing DFL losses, such as the SPO+ surrogate loss~\citep{elmachtoub2022smart}, and thereby turns standard risk-neutral DFL methods into risk-sensitive variants without increasing the complexity of solving at inference time.

To evaluate the framework, we introduce benchmark datasets that expose the trade-off between expected regret and tail regret. Building on the PyEPO framework~\citep{pyepo}, we construct knapsack, portfolio, and shortest-path instances in which candidate solutions exhibit heterogeneous tail behavior: some solutions minimize the expected regret, while others minimize the CVaR of regret.

The contributions of this paper are as follows.
\begin{itemize}
    \item We derive a smooth, entropy-regularized surrogate for the CVaR of regret and show, via Danskin's theorem, that its gradient is the gradient of the underlying DFL loss scaled by a closed-form instance weight. This yields RESTRO, a plug-in reweighting scheme for any surrogate-loss DFL method.
    \item For the SPO+ instantiation, we prove a generalization bound: with high probability, the population CVaR of the true regret is upper-bounded by the empirical RESTRO objective plus a complexity term, and we quantify the bias introduced by the entropy smoothing.
    \item We introduce benchmark datasets with heterogeneous long-tail cost behavior, in which minimizing expected regret and minimizing tail regret lead to different decisions.
    \item In an extensive empirical study on knapsack, portfolio, shortest-path, and energy-aware scheduling problems, we show that RESTRO substantially reduces tail regret on heterogeneous data without sacrificing performance in standard risk-neutral settings. 
\end{itemize}

The remainder of the paper is organized as follows. Section~\ref{sec:related} reviews related literature. Section~\ref{sec:method} presents the methodology, including the smooth CVaR surrogate, its gradient, the training procedure, and the generalization guarantee. Section~\ref{sec:experiments} reports the experimental evaluation, and Section~\ref{sec:conclusion} concludes.

\section{Related Literature}\label{sec:related}

\paragraph{Decision-focused learning.}
Contextual stochastic optimization has received increasing attention in the operations research and ML communities, leading to the predict-then-optimize (PtO) framework. \citet{bertsimas2020predictive} formalized PtO as a two-stage procedure in which a model first predicts uncertain parameters from contextual features and then solves an optimization problem with these estimates. Decision-focused learning (DFL), in contrast, trains the predictive model to optimize downstream decision quality directly. One family of DFL techniques achieves this through surrogate losses~\citep{mulamba2020discrete, mandi22a, elmachtoub2022smart, cave24, berden2025solver, mandi2026feasibilityaware}. Another family differentiates through a smoothed optimization problem, obtained either by regularization~\citep{mandi2020interior, amos2017optnet} or by perturbation~\citep{PogancicPMMR20, berthet2020learning, niepert2021implicit, silvestri2026score}. To compute solutions and their gradients, smoothing-based methods rely on differentiable solvers such as \emph{cvxpylayers}~\citep{agrawal2019differentiable} and \emph{OptNet}~\citep{amos2017optnet}. Approximate approaches such as DYS-Net~\citep{mckenzie2024learning} avoid exact optimization by using projected gradient methods. \citet{Ecai25MandiMBG} recently proposed using differentiable solvers to minimize surrogate losses, and \citet{schutte2024robust} proposed alternative loss functions that make the empirical estimation of regret more robust. Only few methods  tackle directly the minimization of the regret through mathematical programming formulations \cite{bucarey2026decision}. All of these methods, however, target risk-neutral objectives and minimize the average regret.

\paragraph{Risk-sensitive decision-focused learning.}
Several recent works pursue risk-averse DFL by modeling uncertainty explicitly. Gen-DFL~\citep{gendfl} uses generative models to learn a contextual distribution over high-risk regions and improves robustness by sampling at test time. Predict-then-Calibrate~\citep{predictthencalibrate} constructs calibrated uncertainty sets that are used in a distributionally robust optimization (DRO) problem at inference. Differentiable DRO layers~\citep{ma2024differentiable} and end-to-end conditional robust optimization~\citep{chenreddy2024end} learn ambiguity or uncertainty sets and solve a (distributionally) robust optimization problem at test time. In contrast to these approaches, we directly train a point estimator to minimize the CVaR of regret. This enables risk-sensitive decision making with the computational efficiency of deterministic solving at inference time. Section~\ref{subsec:motivation} further explains why minimizing the tail of the regret differs from minimizing the tail of the realized objective value, which is the target of Gen-DFL.

\paragraph{Distributionally robust and long-tailed learning.}
Our approach is also connected to distributionally robust stochastic optimization~\citep{duchi2021learning} for empirical risk minimization in ML. This line of work~\citep{Sagawa, qiattentional, li2021tilted} minimizes the worst-case rather than the average training loss in order to improve generalization to unseen test instances~\citep{fidon, namkoong2016stochastic, duchi2019variance}. Several works in this area, such as Equalization Loss~\citep{equalizationloss} and DropLoss~\citep{droploss}, study reweighting strategies for long-tailed learning~\citep{longtail}. Among these, \emph{re-weighted gradient descent}~\citep{kumar2024stochastic} is most closely related to our approach, as it also reweights the instances of each minibatch during gradient updates.

\section{Methodology}\label{sec:method}

This section first introduces the PtO setting and the SPO+ surrogate loss (Section~\ref{subsec:preliminaries}). Section~\ref{subsec:motivation} explains why we target the tail of the regret rather than the tail of the realized objective value. Sections~\ref{subsec:smooth}--\ref{subsec:training} derive the smooth CVaR surrogate, its gradient, and the resulting training procedure. Section~\ref{subsec:theory} establishes a generalization guarantee for the SPO+ instantiation.

\subsection{Preliminaries: Decision-Focused Learning}\label{subsec:preliminaries}

We consider combinatorial optimization (CO) problems with a linear objective function, of the form
\begin{equation}
    \decisionvar^\star(\cost) \in \argmin_{\decisionvar \in \feasspace} \cost^\top \decisionvar,
    \label{eq:cop}
\end{equation}
where $\cost \in \R^d$ is the cost vector and $\feasspace \subset \R^d$ is the feasible set. We denote the optimal objective value by $v^\star(\cost) = \cost^\top \decisionvar^\star(\cost)$ and write $\decisionvar^\star$ for $\decisionvar^\star(\cost)$ when the argument is clear from context.
In the PtO setting, the cost vector $\cost$ is unknown at decision time. Instead, we observe contextual features $\feature$ and predict the cost vector as $\predcost = f_\theta(\feature)$, where $f_\theta$ is a predictive model, typically a neural network or a linear regressor, with parameters $\theta$. The quality of a prediction $\predcost$ is not measured by how well it approximates $\cost$, but by how good the resulting decision $\decisionvar^\star(\predcost)$ is under the true cost vector $\cost$. This is captured by the \emph{regret}, the excess cost incurred by deciding with the predicted rather than the true cost vector. Given a dataset of feature--cost pairs $\{(\feature_i, \cost_i)\}_{i=1}^N$, the empirical average regret is
\begin{equation}
    \frac{1}{N}\sum_{i=1}^N \mathrm{Regret}(\predcost_i, \cost_i)
    =
    \frac{1}{N}\sum_{i=1}^N \left( \cost_i^\top \decisionvar^\star(\predcost_i) - v^\star(\cost_i) \right) \geq 0 .
    \label{eq:regret}
\end{equation}
Training therefore seeks parameters $\theta$ that minimize the empirical average regret \eqref{eq:regret}. Direct gradient-based minimization of \eqref{eq:regret} is, however, not possible. This is because for CO problems, $\decisionvar^\star(\predcost)$ is a piecewise constant function of $\predcost$. Small changes in the predicted costs typically do not change the optimal solution, so the gradient with respect to $\theta$ is zero almost everywhere.

\paragraph{The SPO+ surrogate loss.}
The non-differentiability of the regret is commonly addressed with surrogate losses that approximate the regret while providing informative (sub)gradients. A widely used surrogate is the SPO+ loss~\citep{elmachtoub2022smart}, which we use as the main instantiation in this paper. It is defined through the perturbed cost vector $2\predcost - \cost$ and the corresponding decision $\decisionvar^\star(2\predcost - \cost)$:
\begin{equation}
    \spoplus(\predcost,\cost)
    = -v^\star(\cost)
      + 2\predcost^\top \decisionvar^\star(\cost)
      - (2\predcost - \cost)^\top \decisionvar^\star(2\predcost - \cost).
    \label{eq:spo}
\end{equation}
A subgradient with respect to the predicted cost is
\begin{equation}
\nabla_{\predcost}\, \spoplus(\predcost,\cost)
= 2\bigl(\decisionvar^\star(\cost) - \decisionvar^\star(2\predcost - \cost)\bigr).
\label{eq:spo_grad}
\end{equation}
An important property of SPO+ is that it upper-bounds the regret pointwise, that is, $\mathrm{Regret}(\predcost,\cost) \leq \spoplus(\predcost,\cost)$ for all $\predcost$ and $\cost$~\citep{elmachtoub2022smart}. We use this property in Section~\ref{subsec:theory}.
Training with SPO+ minimizes the empirical average surrogate loss
\begin{equation}
\mathcal{L}_{\mathrm{Avg}}
= \frac{1}{N}\sum_{i=1}^N
\spoplus(\predcost_i,\cost_i),
\end{equation}
\noindent whose gradient with respect to the model parameters follows from the chain rule:
\begin{equation}
\nabla_\theta \mathcal{L}_{\mathrm{Avg}}
=
\frac{1}{N}\sum_{i=1}^N
\left(\frac{\partial \predcost_i}{\partial \theta}\right)^\top
\nabla_{\predcost_i}
\spoplus(\predcost_i,\cost_i).
\label{eq:subgrad}
\end{equation}
This formulation minimizes the \emph{average} regret over the dataset. In the remainder of this section, we adapt the framework to minimize the CVaR of the regret instead.

\paragraph{Conditional Value-at-Risk.}
For a random loss function $\ell$ and a risk level $\alpha \in (0,1)$, $\CVaR_\alpha(\ell)$ measures the expected loss in the worst $(1-\alpha)$ fraction of outcomes, beyond the $\alpha$-quantile (the Value-at-Risk). Its Rockafellar--Uryasev representation~\citep{rockafellarcvar} is
\begin{equation}
    \CVaR_\alpha(\ell) = \min_{\eta \in \R} \left\{ \eta + \frac{1}{1-\alpha}\, \E[(\ell - \eta)_+] \right\},
    \label{eq:cvar_ru}
\end{equation}
where $(\ell - \eta)_+ = \max(0, \ell - \eta)$. For $N$ observed losses $\underline{\bm{\ell}} = (\ell_1, \ldots, \ell_N)$, replacing the expectation by the sample average gives the empirical CVaR
\begin{equation}
\widehat{\CVaR}_\alpha(\underline{\bm{\ell}})
=
\min_{\eta \in \R}
\left\{
\eta + \frac{1}{(1-\alpha)N} \sum_{i=1}^N \left(\ell_i - \eta\right)_+
\right\}.
\label{eq:cvar_emp}
\end{equation}
A minimizer of \eqref{eq:cvar_emp} is the empirical $\alpha$-quantile of the losses, which we denote by $\eta_\alpha$. When $(1-\alpha)N$ is an integer, $\widehat{\CVaR}_\alpha(\bm{\ell})$ is the average of the $(1-\alpha)N$ largest losses.

In this paper, instead of the average in \eqref{eq:regret}, we aim to minimize the empirical CVaR of regret:
\begin{equation}
\min_{\theta}\;
\widehat{\CVaR}_\alpha\bigl( \mathrm{Regret}(\predcost_1,\cost_1), \ldots, \mathrm{Regret}(\predcost_N,\cost_N) \bigr),
\qquad \predcost_i = f_\theta(\feature_i).
\label{eq:cvar_regret}
\end{equation}
For example, with $\alpha = 0.9$, this is the average regret over the 10\% of instances with the highest regret.
\subsection{Motivation: Tail Regret Versus Tail Objective Value}\label{subsec:motivation}

Our objective differs from robustness objectives that minimize the tail risk of the realized objective value, such as the one targeted by Gen-DFL~\citep{gendfl}. The distinction is subtle but important: such objectives focus on adverse \emph{outcome} scenarios, whereas we focus on adverse \emph{regret} scenarios. The two do not in general lead to the same optimizer.

Let $\cost_1,\ldots,\cost_S$ denote a set of cost scenarios. For a candidate decision $\decisionvar$, the objective value in scenario $s$ is $a_s(\decisionvar) = \cost_s^\top \decisionvar$. A tail-risk objective over outcomes evaluates the tail of $\{a_s(\decisionvar)\}_{s=1}^S$. For instance, if we use CVaR as the risk measure, we will consider the CVaR of the realized objectives across all $S$ scenarios in the following way:
\begin{equation}
\label{eq:cvar_obj_example}
   \widehat{\CVaR}_\alpha\bigl( \cost_1^\top \decisionvar , \ldots,\cost_S^\top \decisionvar   
   \bigr).
\end{equation}
This objective asks whether the decision has a low cost in adverse scenarios. It does not ask whether the decision is close to the best possible decision in each scenario. Regret, in contrast, compares the decision against the scenario-specific optimum $\decisionvar_s^\star = \decisionvar^\star(\cost_s)$. The regret of $\decisionvar$ in scenario $s$ is $r_s(\decisionvar) = \cost_s^\top \decisionvar - \cost_s^\top \decisionvar_s^\star$, and our risk-sensitive objective focuses on the tail of these values:
\begin{equation}
\label{eq:cvar_regret_example}
    \widehat{\CVaR}_{\alpha}\bigl(\cost_1^\top \decisionvar - \cost_1^\top \decisionvar_1^\star , \ldots, \cost_S^\top \decisionvar - \cost_S^\top \decisionvar_S^\star \bigr).
\end{equation}
This objective asks in which scenarios the decision is most suboptimal relative to the best decision available for that scenario. Because the scenarios that form the tail  in \eqref{eq:cvar_regret_example} need not be those that form the tail in \eqref{eq:cvar_obj_example}, minimizing the tail objective value and minimizing the tail regret are generally different problems:
\begin{equation}
    \argmin_{\decisionvar}\,
   \widehat{\CVaR}_\alpha\bigl( \cost_1^\top \decisionvar , \ldots,\cost_S^\top \decisionvar   
   \bigr)
    \neq
    \argmin_{\decisionvar}\,
   \widehat{\CVaR}_{\alpha}\bigl(\cost_1^\top \decisionvar - \cost_1^\top \decisionvar_1^\star , \ldots, \cost_S^\top \decisionvar - \cost_S^\top \decisionvar_S^\star \bigr).
\end{equation}

The distinction matters in practice. In portfolio optimization, a tail-risk objective over realized returns focuses on scenarios in which all assets perform poorly. This can lead to highly conservative decisions, such as investing in only a small subset of assets or avoiding risky assets altogether. Such a decision may perform well in extreme loss scenarios, but it gives up large gains in favorable scenarios, where the scenario-wise optimal portfolio achieves a much higher return. The conservative decision can therefore incur a large regret: it protects against adverse outcomes in an absolute sense, but it does not stay close to the scenario-wise optimal decision.

\paragraph{Illustrative example.}
Consider a shortest-path problem with two routes, A and B, and two scenarios: a normal scenario with probability 0.9 and a rare scenario with probability 0.1. Route A always takes 30 minutes. But Route B takes 20 minutes in the normal scenario and 60 minutes in the rare scenario, for instance due to a railway crossing. Route A therefore has a regret of 10 minutes in the normal scenario and zero in the rare scenario. Route B has zero regret in the normal scenario and 30 minutes in the rare scenario. The $\CVaR_{0.9}$ of the regret is 10 for Route A, whose regret never exceeds 10, and 30 for Route B. The $\CVaR_{0.9}$ of the travel time is 30 for Route A and 60 for Route B. Therefore, both CVaR objectives prefer Route A.

Now suppose we replace Route B with Route B$'$, which takes 40 minutes in the normal scenario but just 5 minutes in the rare scenario, for instance because a tunnel that is rarely open is very fast when available.
In this example, Route A incurs zero regret in the normal scenario and 25 minutes of regret in the rare scenario. Route B$'$, on the other hand, has 10 minutes of regret in the normal scenario and zero regret in the rare scenario. The $\CVaR_{0.9}$ of the travel time prefers Route A to Route B$'$, 30 against 40. The $\CVaR_{0.9}$ of the regret, however, prefers Route B$'$ to Route A, 10 against 25. 
\begin{table}[t]
\centering
\caption{Travel times and regrets (in minutes) in the illustrative example. Bold entries mark the route preferred by each CVaR objective.}
\label{tab:example}
\begin{tabular}{l|cc|cc}
\hline
 & \multicolumn{2}{c|}{Case 1} & \multicolumn{2}{c}{Case 2} \\
 & A & B & A & B$'$ \\
\hline
Travel time, normal scenario ($p = 0.9$) & 30 & 20 & 30 & 40 \\
Travel time, rare scenario ($p = 0.1$)   & 30 & 60 & 30 & 5 \\
Regret, normal scenario                  & 10 & 0  & 0  & 10 \\
Regret, rare scenario                    & 0  & 30 & 25 & 0 \\
\hline
$\CVaR_{0.9}$ of travel time             & \textbf{30} & 60 & \textbf{30} & 40 \\
$\CVaR_{0.9}$ of regret                  & \textbf{10} & 30 & 25 & \textbf{10} \\
\hline
\end{tabular}
\end{table}
 Table~\ref{tab:example} summarizes both cases. Minimizing the CVaR of the objective value and minimizing the CVaR of the regret can therefore lead to different decisions.
\subsection{From scenarios to instances}
In the illustrative example, the CVaR of both regret and travel time is evaluated across two scenarios. In PtO problems, these scenarios correspond to realisations of the cost parameters drawn from their conditional distribution given the contextual features. In practice, however, we observe only one realised scenario for each context. Consider an ambulance dispatcher who chooses a route each morning. The dispatcher observes only the travel times that realized that morning. Alternative travel times under the same context remain unobserved.
Thus, each context in the training set $\{(\feature_i,\cost_i)\}_{i=1}^N$ is paired with a single realised cost vector. Hence we cannot directly compute the empirical CVaR for each context. We can, however, compute the empirical CVaR of regret across the $N$ instances. We will show that this measure yields an upper bound on the average per-context CVaR of regret.

To make this precise, let $(\feature,\cost) \sim \mathcal{D}$ and $R_\theta = \mathrm{Regret}(f_\theta(\feature),\cost)$. The decision $\decisionvar^\star(f_\theta(\feature))$ is a deterministic function of the context, so the model, $f_\theta(\feature)$, defines a decision policy, and $R_\theta$ is the regret this policy incurs. We first formalize these two tail measures and then illustrate their differences.
\begin{equation}
\underbrace{\E_{\feature}\bigl[\CVaR_\alpha(R_\theta \mid \feature)\bigr]}_{\text{per-instance tail}}
\qquad \text{and} \qquad
\underbrace{\CVaR_\alpha(R_\theta)}_{\text{joint-distribution tail}} .
\label{eq:two_tails}
\end{equation}
The \emph{per-instance tail} computes the CVaR of regret under the conditional distribution of $\cost$ given $\feature$, then takes the expectation of this conditional CVaR over the marginal distribution of $\feature$:
\begin{equation}
\E_{\feature}\Bigl[\CVaR_\alpha\bigl(R_\theta \mid \feature\bigr)\Bigr]
= \E_{\feature}\Bigl[\,\min_{\eta \in \R}\Bigl\{\eta + \tfrac{1}{1-\alpha}\,\E_{\cost \mid \feature}\bigl[(R_\theta - \eta)_+\bigr]\Bigr\}\Bigr].
\label{eq:tail_context}
\end{equation}
The \emph{joint-distribution tail} computes the CVaR of regret under the joint distribution of $\feature$ and $\cost$:
\begin{equation}
\CVaR_\alpha\bigl(R_\theta\bigr)
= \min_{\eta \in \R}\Bigl\{\eta + \tfrac{1}{1-\alpha}\,\E_{(\feature,\cost)}\bigl[(R_\theta - \eta)_+\bigr]\Bigr\}.
\label{eq:tail_policy}
\end{equation}
Both expressions follow from the Rockafellar--Uryasev representation~\eqref{eq:cvar_ru}. They differ in the threshold $\eta$. In \eqref{eq:tail_context}, each context has its own threshold, so the tail is formed within each context. In \eqref{eq:tail_policy}, one threshold is formed across contexts. RESTRO computes a single threshold per minibatch and therefore targets the joint distribution tail.
The per-instance tail generalizes the notion of CVaR in the example. The per-instance tail cannot be estimated from one observation per context without modeling the conditional distribution of $\cost$ given $\feature$. Generative and robust approaches take this route~\citep{gendfl,chenreddy2024end}, at the cost of solving a stochastic or robust problem at inference time. The empirical CVaR in \eqref{eq:cvar_regret}, on the other hand, estimates the  CVaR from i.i.d.\ samples drawn from the joint distribution~\citep{soma2020statistical}.

We highlight that for risk-neutral regret, per-instance and joint-distribution measures coincide, because $\E[R_\theta] = \E_{\feature}\bigl[\E[R_\theta \mid \feature]\bigr]$. That is why the distinction between per-instance and joint distribution need not be considered in risk-neutral DFL. However, we have to consider their difference, as we deal with CVaR. 
We next prove that CVaR computed using the joint-distribution yields an upper bound on the average per-context CVaR.
\begin{lemma}\label{lem:tail_order}
For every $\theta$ and every $\alpha \in (0,1)$,
\[
\E_{\feature}\bigl[\CVaR_\alpha(R_\theta \mid \feature)\bigr] \;\leq\; \CVaR_\alpha(R_\theta).
\]
\end{lemma}

\begin{proof}{Proof.}
For a threshold $\eta \in \R$ and a context $\feature$, define
\[
\varphi_\theta(\eta;\feature) \;=\; \eta + \frac{1}{1-\alpha}\,\E_{\cost \mid \feature}\bigl[(R_\theta - \eta)_+\bigr].
\]
Now, for  any $\bar\eta \in \R$, we can write the following using linearity of expectation
\[
\E_{\feature}\bigl[\varphi_\theta(\bar\eta;\feature)\bigr]
\;=\; \bar\eta + \frac{1}{1-\alpha}\,\E_{\feature}\Bigl[\E_{\cost\mid\feature}\bigl[(R_\theta - \bar\eta)_+\bigr]\Bigr]
\;=\; \bar\eta + \frac{1}{1-\alpha}\,\E_{(\feature,\cost) }\bigl[(R_\theta - \bar\eta)_+\bigr].
\]
From the definition of CVaR:
\[
\CVaR_\alpha(R_\theta \mid \feature) \;=\; \min_{\eta \in \R} \varphi_\theta(\eta;\feature) \;\leq\; \varphi_\theta(\bar\eta;\feature).
\]
Taking the expectation over $\feature$ on both sides preserves the inequality:
\[
\E_{\feature}\bigl[\CVaR_\alpha(R_\theta \mid \feature)\bigr] \;\leq\; \E_{\feature}\bigl[\varphi_\theta(\bar\eta;\feature)\bigr].
\]
Note that this is true for all values of $\bar\eta$. So,
\[
\E_{\feature}\bigl[\CVaR_\alpha(R_\theta \mid \feature)\bigr]
\;\leq\; \min_{\bar\eta \in \R}\Bigl\{\bar\eta + \frac{1}{1-\alpha}\,\E_{(\feature,\cost) }\bigl[(R_\theta - \bar\eta)_+\bigr]\Bigr\}
\;=\; \CVaR_\alpha(R_\theta).
\]
\emph{Equality.} Suppose the conditional distribution of $R_\theta \mid \feature$ is the same for every $\feature$. Then $\varphi_\theta(\eta;\feature) = \varphi_\theta(\eta)$ does not depend on $\feature$, and both sides equal $\min_{\eta} \varphi_\theta(\eta)$.
\end{proof}
Lemma~\ref{lem:tail_order} has two implications. First, for every $\theta$, the joint distribution tail upper-bounds the average per-instance tail. The objective \eqref{eq:cvar_regret} estimates the joint distribution tail, so a small value of \eqref{eq:cvar_regret} also implies a small average per-instance tail, without modeling the conditional distribution of $\cost$ given $\feature$. The two tails coincide when the conditional distribution of the regret does not depend on the context. 
Second, when this distribution varies across contexts, the joint distribution tail is \emph{more conservative.} Under a shared threshold, instances from contexts in which the policy incurs high regret exceed the threshold more often. The tail therefore contains both unfavorable realizations within a context and contexts in which the policy performs poorly. For a decision-maker who faces a stream of instances, joint distribution tail is the right metric to minimize. This choice follows standard practice in risk-averse machine learning, where CVaR is computed across training examples rather than over the conditional distribution of the output for each input~\citep{curi2020adaptive,soma2020statistical}.
\subsection{A Smooth Surrogate of the CVaR of Regret}\label{subsec:smooth}
The objective~\eqref{eq:cvar_regret} cannot be minimized directly with gradients, for two reasons. First, each regret term is piecewise constant in $\theta$, as discussed in Section~\ref{subsec:preliminaries}. Second, the empirical CVaR is non-smooth in the losses: the set of instances that form the tail changes abruptly as the losses change.
A surrogate such as SPO+ addresses the first issue by replacing the regret in each term. In this paper, we address the second issue by deriving a smooth surrogate of the CVaR itself.

We first reformulate the empirical CVaR through its dual representation, which introduces instance weights, and then add entropy regularization that smooths the otherwise binary weight assignment. The dual~\citep{ang2018dual} of the empirical CVaR can be expressed in the following form:
\begin{equation}
    \widehat{\CVaR}_\alpha(\bm{\ell}) = \max_{q \in \R^N} \left\{ \sum_{i=1}^N q_i \ell_i \;:\; q_i \geq 0,\ \sum_{i=1}^N q_i = 1,\ q_i \leq \frac{1}{(1-\alpha)N} \right\}.
    \label{eq:cvar_dual}
\end{equation}
Let $K := (1-\alpha)N$. Substituting $q_i = \lambda_i / K$ gives $0 \leq \lambda_i \leq 1$ and $\sum_{i=1}^N \lambda_i = K$, so that
\begin{equation}
    \widehat{\CVaR}_\alpha(\bm{\ell}) = \frac{1}{K} \max_{\lambda \in \R^N} \left\{ \sum_{i=1}^N \lambda_i \ell_i \;:\; 0 \leq \lambda_i \leq 1,\ \sum_{i=1}^N \lambda_i = K \right\}.
    \label{eq:cvar_dual2}
\end{equation}

From this perspective, CVaR is a reweighting of the training instances that has higher weight on high-loss instances and lower weight on low-loss ones. The variable $\lambda_i$ is the weight of instance $i$ in the CVaR objective. The constraints distribute a total weight of $K$ over the dataset, with at most unit weight per instance, and maximizing $\sum_i \lambda_i \ell_i$ concentrates this budget on the instances with the highest losses, effectively selecting the worst $(1-\alpha)$ fraction of samples.
Because a linear objective over a polytope attains its maximum at a vertex, the optimal weights are binary: each instance is either fully selected ($\lambda_i = 1$) or discarded ($\lambda_i = 0$).\endnote{The weights are binary when $K=(1-\alpha)N$ is an integer; otherwise, a single instance may receive a fractional weight.} The resulting objective therefore depends non-smoothly on the losses, since small changes in $\ell_i$ can abruptly change which instances are selected. To obtain a smooth surrogate, we replace this hard selection with soft weights by adding a negative binary-entropy term, which encourages interior solutions $0<\lambda_i<1$:
\begin{equation}
    \max_{0 < \lambda_i < 1} \left\{ \sum_{i=1}^N \lambda_i \ell_i - \frac{1}{\beta} \sum_{i=1}^N \bigl( \lambda_i \log \lambda_i + (1-\lambda_i)\log(1-\lambda_i) \bigr) \;:\; \sum_{i=1}^N \lambda_i = K \right\}.
    \label{eq:cvar_smooth_prob}
\end{equation}
The parameter $\beta > 0$ controls the sharpness of the relaxation. A large $\beta$ yields near-binary instance weights and a solution close to the original CVaR objective, that is, a hard selection of the worst $(1-\alpha)$ fraction of samples. A small $\beta$ induces stronger smoothing and spreads the weight more evenly across samples.

To derive the optimal instance weights, we form the Lagrangian associated with the equality constraint,
\begin{equation}
    \mathbb{L}(\lambda, \eta) = \sum_{i=1}^N \lambda_i \ell_i - \frac{1}{\beta} \sum_{i=1}^N \bigl( \lambda_i \log \lambda_i + (1-\lambda_i)\log(1-\lambda_i) \bigr) - \eta \Bigl( \sum_{i=1}^N \lambda_i - K \Bigr),
    \label{eq:lagrangian}
\end{equation}
and set its derivative with respect to $\lambda_i$ to zero, which characterizes the interior optimum:
\begin{equation}
    \frac{\partial \mathbb{L}}{\partial \lambda_i} = \ell_i - \frac{1}{\beta} \log \frac{\lambda_i}{1-\lambda_i} - \eta = 0 .
    \label{eq:foc}
\end{equation}
Solving for the optimal weight $\lambda_i^\star$ gives
\begin{equation}
    \log \frac{\lambda_i^\star}{1-\lambda_i^\star}
    = \beta(\ell_i - \eta)
    \quad\Longrightarrow\quad
    \lambda_i^\star
    = \frac{1}{1+e^{-\beta(\ell_i - \eta)}}
    = \sigma\bigl(\beta(\ell_i - \eta)\bigr),
    \label{eq:sigmoid}
\end{equation}
where $\sigma(z) = 1/(1+e^{-z})$ is the sigmoid function.

\paragraph{The threshold $\eta$: exact computation and quantile approximation.}
The Lagrange multiplier $\eta$ is determined by the budget constraint, that is, it is the solution $\eta_\beta^\star$ of
\begin{equation}
    \sum_{i=1}^N \sigma\bigl(\beta(\ell_i - \eta)\bigr) = K .
    \label{eq:eta_exact}
\end{equation}
The left-hand side is continuous and strictly decreasing in $\eta$, and it ranges from $N$ to $0$, so \eqref{eq:eta_exact} has a unique solution that can be computed to any precision by bisection. As $\beta \to \infty$, the sigmoid approaches a step function, and $\eta_\beta^\star$ approaches the empirical $\alpha$-quantile $\eta_\alpha$ of the losses, which is the minimizer in \eqref{eq:cvar_emp}. In practice, we therefore approximate the threshold by $\eta_\alpha$, which avoids solving \eqref{eq:eta_exact} in every minibatch. Section~\ref{subsec:exp_exact_eta} shows empirically that both choices lead to nearly identical results and that the approximation is generally faster. With either threshold, the smooth CVaR surrogate reads
\begin{equation}
   \mathcal{L}_{\mathit{CVaR}} = \sum_{i=1}^N \lambda_i^\star \ell_i = \sum_{i=1}^N \sigma\bigl(\beta(\ell_i - \eta)\bigr)\, \ell_i .
   \label{eq:cvar_beta_unnorm}
\end{equation}

\paragraph{Normalization.}
To make the sharpness parameter $\beta$ independent of the absolute scale of the losses, we can normalize the losses by the threshold, which gives
\begin{equation}
   \mathcal{L}_{\mathit{CVaR}} = \sum_{i=1}^N \sigma\!\left(\beta\left(\frac{\ell_i}{\eta} - 1\right)\right) \ell_i .
   \label{eq:cvar_beta}
\end{equation}
Equation~\eqref{eq:cvar_beta} is equivalent to \eqref{eq:cvar_beta_unnorm} with the sharpness rescaled to $\beta/\eta$. It assigns high weight to instances whose loss exceeds the threshold $\eta$ and low weight to those below it, and $\beta$ controls the sharpness of this weighting. Normalization is a practical scaling choice rather than an essential ingredient: Section~\ref{subsec:exp_normalization} shows that it does not materially change the results. Unless stated otherwise, the experiments use the normalized form \eqref{eq:cvar_beta} with the quantile threshold $\eta_\alpha$.

\subsection{Gradient of the Smooth Surrogate}\label{subsec:gradient}

The smooth CVaR objective depends on the model parameters through the instance losses $\ell_i$, which are functions of the predicted costs $\predcost_i$. We now derive its gradient with respect to these predictions, which enables gradient-based training of the predictive model.

Define the smooth CVaR surrogate $G(\predcost)$ as the optimal value of the entropy-regularized problem \eqref{eq:cvar_smooth_prob},
\begin{equation}
G(\predcost) = \max_{\lambda \in \Lambda}\ \psi(\predcost, \lambda),
\qquad
\psi(\predcost, \lambda) = \sum_{i=1}^{N} \lambda_i\, \ell_i(\predcost_i, \cost_i) - \frac{1}{\beta} \sum_{i=1}^{N} \bigl( \lambda_i \log \lambda_i + (1 - \lambda_i) \log(1 - \lambda_i) \bigr),
\label{eq:G_def}
\end{equation}
where $\Lambda = \{\lambda : 0 < \lambda_i < 1,\ \sum_{i=1}^{N} \lambda_i = K\}$ and $\ell_i$ is the instance-level DFL loss (e.g., $\spoplus$). We use Danskin's theorem~\citep{danskin1966theory, bertsekas1997nonlinear}: if $\psi(\predcost, \lambda)$ is concave in $\lambda$ and admits a unique maximizer $\lambda^\star(\predcost)$, then the gradient of the max-function $G$ equals the partial derivative of $\psi$ with respect to $\predcost$, evaluated at the maximizer. The entropy term is strictly concave and the constraint set is convex, so both conditions hold.

To see why the dependence of $\lambda^\star$ on $\predcost$ can be ignored, apply the chain rule:
\begin{equation}
\frac{d G}{d \predcost_i} = \frac{\partial \psi(\predcost, \lambda^\star)}{\partial \predcost_i} + \sum_{j=1}^{N} \frac{\partial \psi(\predcost, \lambda^\star)}{\partial \lambda_j} \cdot \frac{\partial \lambda_j^\star}{\partial \predcost_i}.
\label{eq:chain_rule}
\end{equation}
At the optimum, the first-order conditions of the inner maximization (Equation~\eqref{eq:foc}) make the derivative of the Lagrangian with respect to each $\lambda_j$ vanish. Moreover, since $\sum_j \lambda_j^\star = K$ for all $\predcost$, the multiplier contribution $\eta\sum_j \partial\lambda_j^\star/\partial\predcost_i$ is zero, so the indirect effect (the second term) vanishes. The entropy term does not depend on $\predcost$, so its partial derivative with respect to $\predcost_i$ is also zero. Only the direct effect remains, and
\begin{equation}
\nabla_{\predcost_i} G(\predcost)
=
\lambda_i^\star \cdot
\nabla_{\predcost_i}
\ell_i(\predcost_i, \cost_i).
\label{eq:danskin_grad}
\end{equation}
Hence the optimal weights $\lambda^\star$ can be treated as constants during backpropagation. Substituting the closed-form weights of \eqref{eq:cvar_beta} gives the explicit gradient
\begin{equation}
\nabla_{\predcost_i}
\mathcal{L}_{\mathit{CVaR}}
=
\sigma\!\left(
\beta\left(\frac{\ell_i}{\eta}-1\right)
\right)
\nabla_{\predcost_i}
\ell_i(\predcost_i,\cost_i).
\label{eq:grad_explicit}
\end{equation}

This reveals a key practical advantage: the risk-sensitive gradient is simply the gradient of the loss $\ell_i$ scaled by the instance weight $\lambda_i^\star$. The approach can therefore be applied as a \emph{plug-in} to any existing DFL loss by reweighting the gradients before the parameter update. For example, with the SPO+ surrogate, \eqref{eq:spo_grad} gives
\begin{equation}
\nabla_{\predcost_i}
\mathcal{L}_{\mathit{CVaR}}
=
2\,\lambda_i^\star
\bigl(
\decisionvar^\star(\cost_i) - \decisionvar^\star(2\predcost_i - \cost_i)
\bigr),
\end{equation}
which is the standard SPO+ subgradient scaled by the CVaR-induced instance weight.

\subsection{Risk-Sensitive Decision-Focused Training}\label{subsec:training}

Algorithm~\ref{alg:restro} summarizes the resulting risk-sensitive DFL procedure. The standard DFL pipeline of predicting, optimizing, and computing the DFL loss remains unchanged. Our modification introduces instance-dependent weights that rescale the gradients according to each instance's contribution to the tail of the decision regret.

In practice, we optimize a stochastic approximation of the empirical CVaR objective using minibatches. At each iteration, we compute the CVaR statistics, including the threshold $\eta$, over the current minibatch $\mathcal{B}$ rather than over the full dataset. This yields a minibatch-based stochastic estimate of the CVaR gradient and allows scalable training.
\begin{algorithm}
\caption{Decision-focused learning with risk-sensitive reweighting (RESTRO)}
\label{alg:restro}
\begin{algorithmic}[1]
\Require Training data $\{(\feature_i,\cost_i)\}_{i=1}^N$,
model $f_\theta$, risk level $\alpha$,
sharpness $\beta$, learning rate $\gamma$
\While{not converged}
    \State Sample minibatch $\mathcal B$
    \For{$i \in \mathcal B$}
        \State $\predcost_i \gets f_\theta(\feature_i)$
        \State $\decisionvar_i^\star \gets \argmin_{\decisionvar\in\feasspace} \cost_i^\top \decisionvar$
        \State $\ell_i \gets \ell(\predcost_i,\cost_i)$ \Comment{DFL loss of choice (e.g., SPO+)}
    \EndFor
    \State $\eta \gets
    \mathrm{Quantile}_\alpha(\{\ell_i\}_{i\in\mathcal B})$ \Comment{or solve \eqref{eq:eta_exact} by bisection}
    \State $\lambda_i \gets
    \sigma\big(\beta(\ell_i/\eta-1)\big)$
    for all $i\in\mathcal B$
    \State $g \gets
    \sum_{i\in\mathcal B}
    \lambda_i \nabla_\theta \ell(\predcost_i,\cost_i)$ \Comment{$\lambda_i$ treated as constants}
    \State $\theta \gets \theta-\gamma g$
\EndWhile
\end{algorithmic}
\end{algorithm}
We refer to the resulting framework as \emph{REweighted Subgradients for Tail Regret Optimization} (RESTRO). It can be combined with any decision-focused loss. Our main instantiation uses the SPO+ surrogate and is denoted by $\restro{SPO+}$. 

\subsection{Generalization Guarantee for RESTRO with SPO+}\label{subsec:theory}

Although the reweighting mechanism applies to any DFL loss, we will prove generalization guarantee considering SPO+ as the DFL loss. Our proof rests on a chain of three upper bounds. First, because SPO+ upper-bounds the regret pointwise, the CVaR of the SPO+ loss upper-bounds the CVaR of the regret. Second, the population CVaR of the SPO+ loss is close to its empirical CVaR, uniformly over the hypothesis class. Third, the entropy-regularized objective that RESTRO minimizes upper-bounds the empirical CVaR of the SPO+ loss. $\restro{SPO+}$ therefore minimizes a controlled upper bound on the tail regret. This section makes the argument precise. Surrogates such as IMLE and PFY do not upper-bound the regret, so the generalization guarantee does not extend to them directly.

\paragraph{Notation.}
We write $Z=(\feature,\cost)$ for a feature--cost pair drawn from a distribution $\mathcal{D}$, and $Z_i=(\feature_i,\cost_i)$, $i=1,\ldots,N$, for the training data. The predictive model $f_\theta$ ranges over the hypothesis class $\mathcal{F}=\{f_\theta:\theta\in\Theta\}$, and we write $f$ for a generic element of $\mathcal{F}$. For $f\in\mathcal{F}$, we define the regret and the SPO+ loss of a pair $Z$ as
\[
  r_f(Z)=\mathrm{Regret}\bigl(f(\feature),\cost\bigr),
  \qquad
  \ell_f(Z)=\spoplus\bigl(f(\feature),\cost\bigr),
\]
so that the instance losses of Section~\ref{subsec:smooth} are $\ell_i=\ell_f(Z_i)$. Since this section only concerns SPO+, we omit the superscript on $\ell_f$. For a random variable $L$, $\CVaR_\alpha(L)$ is defined by \eqref{eq:cvar_ru}, and we write $\widehat{\CVaR}_\alpha(\ell_f)$ for the empirical CVaR \eqref{eq:cvar_emp} of the losses $\bigl(\ell_f(Z_1),\ldots,\ell_f(Z_N)\bigr)$. As in Section~\ref{subsec:smooth}, $K=(1-\alpha)N$.

\paragraph{The RESTRO objective.}
Let $H(\lambda)=-\lambda\log\lambda-(1-\lambda)\log(1-\lambda)$ denote the binary entropy, with $H(0)=H(1)=0$. We define the RESTRO objective as
\begin{equation}
  \widehat{R}_{\alpha,\beta}(f)
  =\frac{1}{K}
  \max_{\substack{0\leq\lambda_i\leq1\\ \sum_{i=1}^{N}\lambda_i=K}}
  \left\{
    \sum_{i=1}^{N}\lambda_i\,\ell_f(Z_i)
    +\frac{1}{\beta}\sum_{i=1}^{N}H(\lambda_i)
  \right\}.
  \label{eq:restro_objective}
\end{equation}
The entropy term is the same as the one in \eqref{eq:cvar_smooth_prob}, written with $H$. Closing the box constraints does not change the maximum, because the maximizer \eqref{eq:sigmoid} is interior. Hence $\widehat{R}_{\alpha,\beta}(f)=G(\predcost)/K$, where $G$ is the smooth surrogate \eqref{eq:G_def} with $\ell_i=\ell_f(Z_i)$. By \eqref{eq:danskin_grad}, its gradient is $\frac{1}{K}\sum_{i}\lambda_i^\star\nabla\ell_i$, which is the RESTRO update direction up to the constant factor $1/K$. The guarantee below therefore applies to the objective whose gradient RESTRO follows, computed on the full training set with the exact threshold $\eta_\beta^\star$ and the unnormalized weights \eqref{eq:cvar_beta_unnorm}. The minibatch quantile threshold $\eta_\alpha$ and the normalization \eqref{eq:cvar_beta} are practical approximations of this objective that the bound does not cover. Section~\ref{sec:experiments} shows that they lead to nearly identical results.

\begin{assumption}[Standing assumptions]\label{ass:restro}
\leavevmode
\begin{enumerate}
  \item[(A1)] $Z_1,\ldots,Z_N$ are i.i.d.\ draws from $\mathcal{D}$, and
        $\alpha\in(0,1)$.
  \item[(A2)] The feasible set $\feasspace$ is compact, the cost vector $\cost$ is bounded almost surely, and the hypothesis class
        $\mathcal{F}$ is uniformly bounded. Consequently, there exists
        $U<\infty$ such that
        \[
          0\;\leq\;r_f(Z)\;\leq\;\ell_f(Z)\;\leq\;U
          \qquad\text{almost surely, for every }f\in\mathcal{F}.
        \]
        The middle inequality is the pointwise upper bound of SPO+ on the regret (Section~\ref{subsec:preliminaries} and \citealt{elmachtoub2022smart}).
  \item[(A3)] The sharpness parameter satisfies $\beta>0$.
\end{enumerate}
\end{assumption}

The bound is stated in terms of the complexity of the class of hinge-transformed SPO+ losses that appears in the Rockafellar--Uryasev representation \eqref{eq:cvar_ru}:
\[
  \mathcal{H}
  =\Bigl\{
    Z\mapsto\bigl(\ell_f(Z)-\eta\bigr)_+
    \;:\;f\in\mathcal{F},\ \eta\in[0,U]
  \Bigr\},
\]
where $\eta$ plays the role of the threshold variable in \eqref{eq:cvar_ru}. Let $\mathcal{Z}_N=(Z_1,\ldots,Z_N)$, and let
\[
  \widehat{\mathfrak{R}}_{\mathcal{Z}_N}(\mathcal{H})
  =\E_{\varsigma}
   \Bigl[\sup_{h\in\mathcal{H}}\tfrac{1}{N}\textstyle\sum_{i=1}^{N}\varsigma_i\, h(Z_i)\Bigr],
  \qquad
  \mathfrak{R}_N(\mathcal{H})
  =\E_{\mathcal{Z}_N}\Bigl[\widehat{\mathfrak{R}}_{\mathcal{Z}_N}(\mathcal{H})\Bigr]
\]
denote the empirical and the expected Rademacher complexity of $\mathcal{H}$, where $\varsigma_1,\ldots,\varsigma_N$ are i.i.d.\ Rademacher variables.

\begin{proposition}[Generalization bound for RESTRO with SPO+]\label{prop:restro-generalization}
Let Assumption~\ref{ass:restro} hold, and let $\delta\in(0,1)$. Then, with probability at least $1-\delta$, simultaneously for every $f\in\mathcal{F}$,
\[
  \CVaR_{\alpha}\bigl(r_f(Z)\bigr)
  \;\leq\;
  \widehat{R}_{\alpha,\beta}(f)+\epsilon_N(\delta),
  \qquad
  \epsilon_N(\delta)
  =\frac{1}{1-\alpha}
   \left(
     2\mathfrak{R}_N(\mathcal{H})
     +U\sqrt{\frac{\log(1/\delta)}{2N}}
   \right).
\]
\end{proposition}

\begin{proof}{Proof.}
\emph{Step 1: the entropy term is slack.}
Since $H(\lambda_i)\geq0$ on $[0,1]$ and $\beta>0$, every feasible $\lambda$ in \eqref{eq:restro_objective} satisfies
$\sum_i\lambda_i\ell_f(Z_i)
 \leq\sum_i\lambda_i\ell_f(Z_i)
      +\tfrac{1}{\beta}\sum_iH(\lambda_i)$.
Taking the maximum over the feasible set, dividing by $K$, and using the dual representation \eqref{eq:cvar_dual2} of the empirical CVaR gives
\begin{equation}
  \widehat{\CVaR}_{\alpha}(\ell_f)
  \;\leq\;\widehat{R}_{\alpha,\beta}(f).
  \label{eq:entropy-slack}
\end{equation}

\emph{Step 2: Rockafellar--Uryasev representation on a bounded range.}
For a random variable $L$, let
$\phi(\eta)=\eta+\tfrac{1}{1-\alpha}\E\bigl[(L-\eta)_+\bigr]$,
so that $\CVaR_\alpha(L)=\min_{\eta\in\R}\phi(\eta)$ by \eqref{eq:cvar_ru}.
If $L\in[0,U]$ almost surely, the minimum is attained on $[0,U]$. For
$\eta<0$, we have $(L-\eta)_+=L-\eta$, hence
$\phi(\eta)=\tfrac{\E[L]}{1-\alpha}
            +\eta\bigl(1-\tfrac{1}{1-\alpha}\bigr)$,
which is nonincreasing in $\eta$ because $\tfrac{1}{1-\alpha}>1$. For
$\eta>U$, we have $(L-\eta)_+=0$, hence $\phi(\eta)=\eta>U=\phi(U)$.
Applying this to $\ell_f(Z)$ under $\mathcal{D}$ and to the empirical distribution of $\ell_f(Z_1),\ldots,\ell_f(Z_N)$ gives
\[
  \CVaR_{\alpha}\bigl(\ell_f(Z)\bigr)
  =\min_{\eta\in[0,U]}
   \left\{\eta+\frac{1}{1-\alpha}
   \E\bigl[(\ell_f(Z)-\eta)_+\bigr]\right\},
\]
\[
  \widehat{\CVaR}_{\alpha}(\ell_f)
  =\min_{\eta\in[0,U]}
   \left\{\eta+\frac{1}{(1-\alpha)N}
   \sum_{i=1}^{N}(\ell_f(Z_i)-\eta)_+\right\}.
\]

\emph{Step 3: one-sided comparison of minima.}
For any functions $a$ and $b$, if $\eta_b$ attains $\min_\eta b(\eta)$, then
$\min_\eta a(\eta)\leq a(\eta_b)
  =b(\eta_b)+\bigl(a(\eta_b)-b(\eta_b)\bigr)
  \leq\min_\eta b(\eta)+\sup_\eta\bigl(a(\eta)-b(\eta)\bigr)$.
Taking $a$ and $b$ as the objectives of the two displays in Step~2, the $\eta$ terms cancel, and
\begin{equation}
  \CVaR_{\alpha}\bigl(\ell_f(Z)\bigr)
  -\widehat{\CVaR}_{\alpha}(\ell_f)
  \leq\frac{1}{1-\alpha}\,
      \sup_{\eta\in[0,U]}
      \left\{
        \E\bigl[h_{f,\eta}(Z)\bigr]
        -\frac{1}{N}\sum_{i=1}^{N}h_{f,\eta}(Z_i)
      \right\},
  \label{eq:min-compare}
\end{equation}
where $h_{f,\eta}(Z)=(\ell_f(Z)-\eta)_+\in\mathcal{H}$.

\emph{Step 4: uniform convergence over $\mathcal{H}$.}
The supremum in \eqref{eq:min-compare} is bounded by the supremum over
all of $\mathcal{H}$, which ranges over $f$ and $\eta$ jointly. Every
$h\in\mathcal{H}$ takes values in $[0,U]$, so replacing a single sample
changes $\Phi(\mathcal{Z}_N)=\sup_{h\in\mathcal{H}}
\bigl(\E[h(Z)]-\tfrac1N\sum_i h(Z_i)\bigr)$ by at most $U/N$. By
McDiarmid's inequality and symmetrization, with probability at least
$1-\delta$,
\begin{equation}
  \sup_{h\in\mathcal{H}}
  \left\{\E[h(Z)]-\frac{1}{N}\sum_{i=1}^{N}h(Z_i)\right\}
  \;\leq\;
  2\mathfrak{R}_N(\mathcal{H})+U\sqrt{\frac{\log(1/\delta)}{2N}}.
  \label{eq:unif-conv}
\end{equation}
Combining \eqref{eq:min-compare} and \eqref{eq:unif-conv} gives, with
probability at least $1-\delta$ and simultaneously for all
$f\in\mathcal{F}$,
\begin{equation}
  \CVaR_{\alpha}\bigl(\ell_f(Z)\bigr)
  \leq\widehat{\CVaR}_{\alpha}(\ell_f)
       +\epsilon_N(\delta).
  \label{eq:cvar-gen}
\end{equation}

\emph{Step 5: conclusion.}
Chaining \eqref{eq:entropy-slack} into \eqref{eq:cvar-gen} gives
$\CVaR_{\alpha}\bigl(\ell_f(Z)\bigr)
  \leq\widehat{R}_{\alpha,\beta}(f)+\epsilon_N(\delta)$.
By (A2), $r_f(Z)\leq\ell_f(Z)$ almost surely, and
$\CVaR_\alpha$ is monotone, so
$\CVaR_{\alpha}\bigl(r_f(Z)\bigr)
 \leq\CVaR_{\alpha}\bigl(\ell_f(Z)\bigr)$, which yields
the claim.\Halmos
\end{proof}

The bound in Proposition~\ref{prop:restro-generalization} involves the expected Rademacher complexity $\mathfrak{R}_N(\mathcal{H})$, which cannot be computed from the data. The following remark gives a fully data-dependent version.

\begin{remark}[Data-dependent version]\label{rem:empirical-rademacher}
Replacing one sample point changes $\widehat{\mathfrak{R}}_{\mathcal{Z}_N}(\mathcal{H})$
by at most $U/N$, since every $h\in\mathcal{H}$ takes values in $[0,U]$.
Applying McDiarmid's inequality to the lower tail, which excludes the event that the empirical complexity is unusually small, gives
\[
  \Pr\Bigl[
    \widehat{\mathfrak{R}}_{\mathcal{Z}_N}(\mathcal{H})
    -\mathfrak{R}_N(\mathcal{H})
    \leq-t
  \Bigr]
  \;\leq\;\exp\Bigl(-\frac{2t^2N}{U^2}\Bigr).
\]
Setting the right-hand side to $\delta/2$ gives, with probability at
least $1-\delta/2$,
\[
  \mathfrak{R}_N(\mathcal{H})
  \;\leq\;\widehat{\mathfrak{R}}_{\mathcal{Z}_N}(\mathcal{H})
          +U\sqrt{\frac{\log(2/\delta)}{2N}}.
\]
Applying \eqref{eq:unif-conv} with $\delta/2$ in place of $\delta$ and taking a union bound, with probability at least $1-\delta$,
\begin{equation}
  \sup_{h\in\mathcal{H}}
  \left\{\E[h(Z)]-\frac{1}{N}\sum_{i=1}^{N}h(Z_i)\right\}
  \;\leq\;
  2 \widehat{\mathfrak{R}}_{\mathcal{Z}_N}(\mathcal{H})
 + 3 U\sqrt{\frac{\log(2/\delta)}{2N}} .
\end{equation}
Proposition~\ref{prop:restro-generalization} therefore also holds with
\[
  \epsilon_N(\delta)
  =\frac{1}{1-\alpha}
   \left(
     2\widehat{\mathfrak{R}}_{\mathcal{Z}_N}(\mathcal{H})
     +3U\sqrt{\frac{\log(2/\delta)}{2N}}
   \right).
\]
\end{remark}

Inequality~\eqref{eq:entropy-slack} is one-sided, since the entropy term can only increase the objective. The next remark quantifies how much the smoothing can overestimate the empirical CVaR, and how the sharpness parameter $\beta$ should scale with $N$ to keep this bias below the statistical error.

\begin{remark}[Smoothing bias of the entropy term]\label{rem:entropy-bias}
The gap in \eqref{eq:entropy-slack} is controlled by $\beta$ rather than by $N$. By concavity of $H$ and $\sum_i\lambda_i=K$, Jensen's
inequality gives $\sum_{i=1}^{N}H(\lambda_i)\leq N\,H(K/N)=N\,H(1-\alpha)$,
hence
\[
  \widehat{\CVaR}_{\alpha}(\ell_f)
  \;\leq\;\widehat{R}_{\alpha,\beta}(f)
  \;\leq\;\widehat{\CVaR}_{\alpha}(\ell_f)
          +\frac{H(1-\alpha)}{\beta\,(1-\alpha)}.
\]
The smoothing bias is therefore at most of order $1/\beta$, and this bound does not improve with $N$. Keeping it at or below the statistical rate $\epsilon_N(\delta)=O(N^{-1/2})$ requires $\beta=\Omega(\sqrt{N})$. With this scaling, the RESTRO objective is a consistent surrogate for the empirical CVaR.
\end{remark}

\section{Experimental Evaluation}\label{sec:experiments}

This section evaluates RESTRO empirically. Section~\ref{subsec:dataset} describes the benchmark problems and our heterogeneous long-tail data generation, and Section~\ref{subsec:setup} the experimental setup. Section~\ref{subsec:main_results} reports the main results. The remaining subsections compare the exact and approximate thresholds (Section~\ref{subsec:exp_exact_eta}), and an energy-aware scheduling problem (Section~\ref{subsec:exp_energy}), and study the effect of normalization (Section~\ref{subsec:exp_normalization}).

\subsection{Synthetic Cost Generation with Heterogeneous Long-Tail Perturbations}\label{subsec:dataset}

We build on \texttt{PyEPO}~\citep{pyepo}, a widely used benchmark for PtO problems. \texttt{PyEPO} generates cost vectors for CO problems synthetically through a nonlinear mapping of contextual features. In the shortest-path setting, for example, the cost of edge $j$ in instance $i$ is generated as
\begin{equation}
\upsilon_{ij}
=
\frac{\left(\frac{1}{\sqrt{p}}(B\feature_i)_j + 3\right)^{\mathrm{deg}} + 1}{3.5^{\mathrm{deg}}},
\qquad
c_{ij}
=
\upsilon_{ij}
\cdot \epsilon_{ij},
\label{eq:randomcost}
\end{equation}
where $\feature_i \in \R^p$ is the contextual feature vector of instance $i$, $B \in \{0,1\}^{d \times p}$ is a random binary matrix shared across all instances, and $\epsilon_{ij}$ is a multiplicative noise term sampled independently for each cost coefficient. In this construction, all costs are generated with the same transformation and noise model, so the induced distributions of the cost coefficients share the same parametric form and differ mainly in scale. As a result, \emph{the expectation and the CVaR tend to be positively correlated}, and \emph{decisions optimized for average performance often remain optimal with respect to tail risk}. To create a genuine mismatch between average performance and tail risk, we introduce datasets in which the decisions that are optimal on average differ from those that are optimal under tail-risk criteria.

\paragraph{Tail-aware synthetic data generation.}
We generate datasets for three CO problems: shortest path, knapsack, and portfolio optimization. For all three, we modify the \texttt{PyEPO} generators to produce cost vectors with \emph{heterogeneous long-tail behavior}. The modification is heterogeneous because it applies only to a subset $\mathcal{S}$ of the cost coefficients: the coefficients in $\mathcal{S}$ follow long-tailed distributions, while the others follow the usual \texttt{PyEPO} distributions. The aim is to create coefficients that look attractive under average performance but become undesirable under tail-risk criteria. This setting mimics real-world environments in which only some assets, routes, or items are exposed to rare but severe disruptions, while the others show regular variability.

\paragraph{Shortest path.}
We first generate $\upsilon_{ij}$ with the standard \texttt{PyEPO} nonlinear transformation \eqref{eq:randomcost}. For the special edges $j \in \mathcal{S}$, we rescale the cost as
\begin{equation}
\upsilon_{ij} \leftarrow \upsilon_{ij} \cdot s_i \xi_i,
\qquad j \in \mathcal{S},
\end{equation}
where $s_i = 0.1$ with probability $0.8$ and $s_i = 3$ with probability $0.2$, and $\xi_i \sim \mathcal{N}(1,0.01^2)$. Since $\E[s_i] = 0.8 \cdot 0.1 + 0.2 \cdot 3 = 0.68$, these edges have a lower average cost but can incur large costs in the tail. We then apply the \texttt{PyEPO} multiplicative noise to all edges.

\paragraph{Knapsack.}
For knapsack, we use the \texttt{PyEPO} construction of the base item values and rescale the values of the special items as in the shortest-path problem, with $s_i = 5.0$ with probability $0.8$, $s_i = 0.05$ with probability $0.2$, and $\xi_i \sim \mathcal{N}(1,0.01^2)$. The special items are taken from the high-weight end of the item ordering, so they are resource intensive. Since $\E[s_i] = 0.8 \cdot 5.0 + 0.2 \cdot 0.05 = 4.01$, these items are profitable on average, but in 20\% of the cases they contribute little value while still consuming substantial capacity. We then apply the \texttt{PyEPO} nonlinear rescaling and multiplicative noise.

\paragraph{Portfolio.}
For portfolio optimization, the base mean returns are computed with the \texttt{PyEPO} mapping
\begin{equation}
\mu_{ij}
=
\left(
0.05\, \frac{(B \feature_i)_j}{\sqrt{p}}
+
0.1^{1/\mathrm{deg}}
\right)^{\mathrm{deg}} .
\end{equation}
For the special assets $j \in \mathcal{S}$, we rescale the base mean return as $\mu_{ij} \leftarrow \mu_{ij} \cdot s_{ij}\xi_{ij}$, where $s_{ij} = 2.0$ with probability $0.8$, $s_{ij} = -1.0$ with probability $0.2$, and $\xi_{ij} \sim \mathcal{N}(1,0.01^2)$. Since $\E[s_{ij}] = 0.8 \cdot 2.0 + 0.2 \cdot (-1.0) = 1.4$, these assets have higher expected returns but can generate losses in the adverse regime. The final returns add factor-driven and idiosyncratic noise, $r_i = \mu_i + L f_i + 0.01 \cdot \sigma \epsilon_i$, where $f_i \sim \mathcal{N}(0,I_p)$, $\epsilon_i \sim \mathcal{N}(0,I_m)$, and $\sigma$ is the noise level.

\subsection{Experimental Setup}\label{subsec:setup}

We evaluate the RESTRO framework instantiated with the SPO+ loss, denoted $\restro{SPO+}$, on portfolio optimization with 50 assets, shortest path on a $5\times5$ grid, and knapsack with 50 items. We compare against a prediction-focused baseline trained with MSE, standard SPO+~\citep{elmachtoub2022smart}. We report results on the standard \texttt{PyEPO} datasets with homogeneous cost vectors and on our heterogeneous long-tail datasets.

For each problem, we use 800 training instances and 200 validation instances to tune the method-specific hyperparameters: the learning rate, the minibatch size, and, for $\restro{SPO+}$, the sharpness $\beta$. Table~\ref{tab:hyperparams} lists the selected hyperparameters. After hyperparameter selection, we evaluate on 1000 test instances. We compute the regret of each test instance and report the CVaR with $\alpha=0.9$ over the 1000 regret values. We repeat this procedure for 5 random seeds, with independently generated training and test instances for each seed, and report the mean and standard deviation of the CVaR across the seeds.

The ML models are implemented in PyTorch~\citep{paszke2019pytorch}, and the CO problems are solved with Gurobi~\citep{gurobi}. The experiments were run on an Intel i7-13800H CPU with 14 cores and 20 threads, and 32\,GB of RAM.

\begin{table}
	\TABLE
	{Selected Hyperparameters for Each Problem, Dataset, and Method.\label{tab:hyperparams}}
	{\begin{tabular}{@{}lllccc@{}}
			\toprule
			Problem & Data & Method & Batch size & $\beta$ & Learning rate \\
			\midrule
			\multirow{8}*{Portfolio (50 assets)}
			& Heterogeneous & MSE & 32 & -- & 0.005 \\
			& Heterogeneous & SPO+ & 128 & -- & 0.005 \\
			& Heterogeneous & $\restro{SPO+}$ (exact $\eta$) & 128 & 50 & 0.005 \\
			& Heterogeneous & $\restro{SPO+}$ (quantile $\eta$) & 128 & 10 & 0.005 \\
			& Homogeneous & MSE & 128 & -- & 0.005 \\
			& Homogeneous & SPO+ & 64 & -- & 0.005 \\
			& Homogeneous & $\restro{SPO+}$ (exact $\eta$) & 64 & 1 & 0.005 \\
			& Homogeneous & $\restro{SPO+}$ (quantile $\eta$) & 64 & 0.1 & 0.005 \\
			\midrule
			\multirow{8}*{Shortest path ($5\times5$ grid)}
			& Heterogeneous & MSE & 128 & -- & 0.005 \\
			& Heterogeneous & SPO+ & 128 & -- & 0.05 \\
			& Heterogeneous & $\restro{SPO+}$ (exact $\eta$) & 64 & 10 & 0.005 \\
			& Heterogeneous & $\restro{SPO+}$ (quantile $\eta$) & 64 & 50 & 0.005 \\
			& Homogeneous & MSE & 64 & -- & 0.005 \\
			& Homogeneous & SPO+ & 128 & -- & 0.005 \\
			& Homogeneous & $\restro{SPO+}$ (exact $\eta$) & 128 & 0.1 & 0.005 \\
			& Homogeneous & $\restro{SPO+}$ (quantile $\eta$) & 128 & 0.1 & 0.005 \\
			\midrule
			\multirow{8}*{Knapsack (50 items)}
			& Heterogeneous & MSE & 128 & -- & 0.05 \\
			& Heterogeneous & SPO+ & 64 & -- & 0.05 \\
			& Heterogeneous & $\restro{SPO+}$ (exact $\eta$) & 32 & 50 & 0.005 \\
			& Heterogeneous & $\restro{SPO+}$ (quantile $\eta$) & 32 & 10 & 0.005 \\
			& Homogeneous & MSE & 64 & -- & 0.005 \\
			& Homogeneous & SPO+ & 64 & -- & 0.05 \\
			& Homogeneous & $\restro{SPO+}$ (exact $\eta$) & 32 & 0.1 & 0.05 \\
			& Homogeneous & $\restro{SPO+}$ (quantile $\eta$) & 128 & 1 & 0.05 \\
			\bottomrule
	\end{tabular}}
	{Hyperparameters were selected on 200 validation instances after training on 800 instances.}
\end{table}

\subsection{Experimental Results with homogeneous and  heterogeneous datasets}\label{subsec:main_results}

Table~\ref{tab:result_homo_main} reports the results on the standard \texttt{PyEPO} datasets with homogeneous cost vectors, and Table~\ref{tab:result_hetero_main} the results on our heterogeneous long-tail datasets.

\paragraph{Performance on homogeneous cost vectors.}
On the homogeneous datasets, SPO+ already achieves a very low CVaR of regret, which leaves little room for improvement. This is expected: on these datasets, average performance and tail risk are closely aligned, so decisions that are near-optimal in expectation also tend to perform well under the CVaR criterion. Consistently, hyperparameter tuning typically selected small values of $\beta$, which produce nearly uniform instance weights and make RESTRO behave similarly to SPO+.

\paragraph{Performance on heterogeneous cost vectors.}
On the heterogeneous benchmarks (Table~\ref{tab:result_hetero_main}), $\restro{SPO+}$ achieves the lowest CVaR of regret on shortest path and knapsack, and it is competitive with the best method on portfolio. Because SPO+ weights all instances equally, it achieves a low mean regret but performs poorly on the long-tail instances with high regret. We highlight that RESTRO achieves very low CVaR of regret by assigning more weight to instances that induce high regret. These results support subgradient reweighting as an effective way to minimize the CVaR of regret.

\begin{table}
	\TABLE
	{$\text{CVaR}_{0.9}$ of Regret on the \texttt{PyEPO} Datasets with Homogeneous Cost Vectors.\label{tab:result_homo_main}}
	{\begin{tabular}{@{}lcccc@{}}
			\toprule
			Problem & MSE & SPO+ & $\restro{SPO+}$ (exact $\eta$) & $\restro{SPO+}$ (quantile $\eta$) \\
			\midrule
			Portfolio & $0.00 \pm 0.00$ & $0.00 \pm 0.00$ & $0.00 \pm 0.00$ & $0.00 \pm 0.00$ \\
			Shortest path & $0.53 \pm 0.11$ & $0.38 \pm 0.07$ & $0.38 \pm 0.06$ & $0.39 \pm 0.07$ \\
			Knapsack & $2.30 \pm 0.30$ & $1.61 \pm 0.07$ & $1.61 \pm 0.11$ & $1.62 \pm 0.09$ \\
			\bottomrule
	\end{tabular}}
	{Mean $\pm$ standard deviation over 5 random seeds; each value is the CVaR over 1000 test instances.}
\end{table}

\begin{table}
	\TABLE
	{$\text{CVaR}_{0.9}$ of Regret on the Datasets with Heterogeneous Long-Tail Cost Vectors.\label{tab:result_hetero_main}}
	{\begin{tabular}{@{}lcccc@{}}
			\toprule
			Problem & MSE & SPO+ & $\restro{SPO+}$ (exact $\eta$) & $\restro{SPO+}$ (quantile $\eta$) \\
			\midrule
			Portfolio & $0.43 \pm 0.01$ & $0.42 \pm 0.01$ & $0.39 \pm 0.01$ & $0.39 \pm 0.01$ \\
			Shortest path & $5.36 \pm 0.83$ & $11.04 \pm 0.96$ & $4.99 \pm 0.63$ & $5.07 \pm 0.55$ \\
			Knapsack & $10.73 \pm 1.50$ & $20.54 \pm 5.11$ & $8.18 \pm 0.84$ & $8.07 \pm 0.90$ \\
			\bottomrule
	\end{tabular}}
	{Mean $\pm$ standard deviation over 5 random seeds; each value is the CVaR over 1000 test instances.}
\end{table}

\subsection{Exact Versus Approximate Threshold}\label{subsec:exp_exact_eta}

In the main experiments, the threshold $\eta$ is approximated by the empirical $\alpha$-quantile $\eta_\alpha$ of the minibatch losses. Here we compare this approximation with the exact threshold $\eta_\beta^\star$, obtained by solving \eqref{eq:eta_exact} by bisection in every minibatch. Table~\ref{tab:exact_eta} reports the $\text{CVaR}_{0.9}$ of regret and the training time for both variants, together with MSE and SPO+.

The exact threshold and the quantile approximation produce nearly identical $\text{CVaR}_{0.9}$ values on all three problems and in both data regimes. The small differences are within the reported variability and do not change any conclusion. The quantile approximation is also generally faster, because it avoids solving the threshold equation by bisection in every minibatch. The approximation used in the main experiments thus preserves the performance of the exact formulation while reducing its computational cost.

\begin{table}
	\TABLE
	{Exact Threshold $\eta_\beta^\star$ Versus Quantile Approximation $\eta_\alpha$: $\text{CVaR}_{0.9}$ of Regret and Training Time (s).\label{tab:exact_eta}}
	{\begin{tabular}{@{}llcccc@{}}
			\toprule
			& & \multicolumn{2}{c}{Heterogeneous} & \multicolumn{2}{c}{Homogeneous} \\
			\cmidrule(lr){3-4}\cmidrule(l){5-6}
			Problem & Method & $\text{CVaR}_{0.9}$ & Time (s) & $\text{CVaR}_{0.9}$ & Time (s) \\
			\midrule
			\multirow{4}*{Shortest path}
			& MSE & $5.36 \pm 0.83$ & $2.11 \pm 0.04$ & $0.53 \pm 0.11$ & $2.28 \pm 0.07$ \\
			& SPO+ & $11.04 \pm 0.96$ & $10.09 \pm 0.33$ & $0.38 \pm 0.07$ & $10.54 \pm 0.31$ \\
			& $\restro{SPO+}$ (exact $\eta$) & $4.99 \pm 0.63$ & $11.41 \pm 0.85$ & $0.38 \pm 0.06$ & $11.67 \pm 0.85$ \\
			& $\restro{SPO+}$ (quantile $\eta$) & $5.07 \pm 0.55$ & $10.37 \pm 0.17$ & $0.39 \pm 0.07$ & $10.11 \pm 0.20$ \\
			\midrule
			\multirow{4}*{Knapsack}
			& MSE & $10.73 \pm 1.50$ & $7.38 \pm 0.78$ & $2.30 \pm 0.30$ & $7.67 \pm 0.53$ \\
			& SPO+ & $20.54 \pm 5.11$ & $37.41 \pm 6.47$ & $1.61 \pm 0.07$ & $30.13 \pm 2.55$ \\
			& $\restro{SPO+}$ (exact $\eta$) & $8.18 \pm 0.84$ & $29.38 \pm 4.16$ & $1.61 \pm 0.11$ & $38.22 \pm 3.58$ \\
			& $\restro{SPO+}$ (quantile $\eta$) & $8.07 \pm 0.90$ & $26.04 \pm 0.85$ & $1.62 \pm 0.09$ & $27.41 \pm 1.88$ \\
			\midrule
			\multirow{4}*{Portfolio}
			& MSE & $0.43 \pm 0.01$ & $2.76 \pm 0.10$ & $0.00 \pm 0.00$ & $8.28 \pm 0.18$ \\
			& SPO+ & $0.42 \pm 0.01$ & $11.21 \pm 0.12$ & $0.00 \pm 0.00$ & $66.10 \pm 0.92$ \\
			& $\restro{SPO+}$ (exact $\eta$) & $0.39 \pm 0.01$ & $12.02 \pm 0.56$ & $0.00 \pm 0.00$ & $69.22 \pm 2.56$ \\
			& $\restro{SPO+}$ (quantile $\eta$) & $0.39 \pm 0.01$ & $11.26 \pm 0.15$ & $0.00 \pm 0.00$ & $65.48 \pm 1.66$ \\
			\bottomrule
	\end{tabular}}
	{Mean $\pm$ standard deviation over 5 random seeds. Shortest path on a $5\times5$ grid, knapsack with 50 items, portfolio with 50 assets. The exact variant uses the unnormalized weights \eqref{eq:cvar_beta_unnorm} and the quantile variant the normalized weights \eqref{eq:cvar_beta}; Table~\ref{tab:normalization} reports all four combinations.}
\end{table}

\subsection{Effect of Normalization}\label{subsec:exp_normalization}

The main experiments normalize the losses by the threshold, as in \eqref{eq:cvar_beta}, to control the scale of the sigmoid transition and reduce the dependence of $\beta$ on the absolute magnitude of the losses. We test whether this normalization materially affects the results by comparing the normalized form \eqref{eq:cvar_beta} with the unnormalized form \eqref{eq:cvar_beta_unnorm}, each with the exact threshold and with the quantile approximation. Table~\ref{tab:normalization} shows that normalization does not materially change the $\text{CVaR}_{0.9}$ values in any of the tested settings, and the small runtime differences are not systematic. The empirical gains of RESTRO therefore come from the tail-aware reweighting rather than from the normalization, which is a practical scaling choice.

\begin{table}
	\TABLE
	{Effect of Normalizing the Losses by the Threshold: $\text{CVaR}_{0.9}$ of Regret and Training Time (s) of $\restro{SPO+}$.\label{tab:normalization}}
	{\begin{tabular}{@{}lllcccc@{}}
			\toprule
			& & & \multicolumn{2}{c}{Heterogeneous} & \multicolumn{2}{c}{Homogeneous} \\
			\cmidrule(lr){4-5}\cmidrule(l){6-7}
			Problem & $\eta$ & Normalized & $\text{CVaR}_{0.9}$ & Time (s) & $\text{CVaR}_{0.9}$ & Time (s) \\
			\midrule
			\multirow{4}*{Shortest path}
			& Exact & No & $4.99 \pm 0.63$ & $11.41 \pm 0.85$ & $0.38 \pm 0.06$ & $11.67 \pm 0.85$ \\
			& Exact & Yes & $5.02 \pm 0.56$ & $10.67 \pm 0.05$ & $0.39 \pm 0.07$ & $10.27 \pm 0.06$ \\
			& Quantile & No & $4.99 \pm 0.42$ & $10.08 \pm 0.09$ & $0.39 \pm 0.07$ & $10.12 \pm 0.07$ \\
			& Quantile & Yes & $5.07 \pm 0.55$ & $10.37 \pm 0.17$ & $0.39 \pm 0.07$ & $10.11 \pm 0.20$ \\
			\midrule
			\multirow{4}*{Knapsack}
			& Exact & No & $8.18 \pm 0.84$ & $29.38 \pm 4.16$ & $1.61 \pm 0.11$ & $38.22 \pm 3.58$ \\
			& Exact & Yes & $8.11 \pm 0.90$ & $27.15 \pm 0.81$ & $1.60 \pm 0.09$ & $35.05 \pm 2.32$ \\
			& Quantile & No & $8.12 \pm 0.82$ & $24.78 \pm 0.91$ & $1.60 \pm 0.11$ & $30.29 \pm 2.21$ \\
			& Quantile & Yes & $8.07 \pm 0.90$ & $26.04 \pm 0.85$ & $1.62 \pm 0.09$ & $27.41 \pm 1.88$ \\
			\midrule
			\multirow{4}*{Portfolio}
			& Exact & No & $0.39 \pm 0.01$ & $12.02 \pm 0.56$ & $0.00 \pm 0.00$ & $69.22 \pm 2.56$ \\
			& Exact & Yes & $0.39 \pm 0.00$ & $11.32 \pm 0.04$ & $0.00 \pm 0.00$ & $64.60 \pm 0.18$ \\
			& Quantile & No & $0.39 \pm 0.00$ & $11.06 \pm 0.09$ & $0.00 \pm 0.00$ & $63.47 \pm 0.17$ \\
			& Quantile & Yes & $0.39 \pm 0.01$ & $11.26 \pm 0.15$ & $0.00 \pm 0.00$ & $65.48 \pm 1.66$ \\
			\bottomrule
	\end{tabular}}
	{Mean $\pm$ standard deviation over 5 random seeds.}
\end{table}
\subsection{Energy-Aware Scheduling}\label{subsec:exp_energy}

Finally, to test RESTRO beyond the \texttt{PyEPO} generators, we evaluate it on the energy-cost-aware scheduling benchmark used in the DFL literature~\citep{DFLsurvey}, with an instance of 10 tasks. The cost parameters are electricity prices that must be predicted from historical features. We consider the original data and a heterogeneous variant with long-tail perturbations. Table~\ref{tab:energy} reports the results.

On the original data, all DFL methods improve over MSE, and RESTRO slightly improves over SPO+. On the heterogeneous variant whereas RESTRO reduces the CVaR of regret below both MSE and SPO+. We have observed the same behavior on the synthetic heterogeneous datasets.
\begin{table}
	\TABLE
	{Energy-Aware Scheduling with 10 Tasks: $\text{CVaR}_{0.9}$ of Regret and Training Time (s).\label{tab:energy}}
	{\begin{tabular}{@{}llcc@{}}
			\toprule
			Data & Method & $\text{CVaR}_{0.9}$ & Time (s) \\
			\midrule
			\multirow{5}*{Original}
			& MSE & 166{,}331.07 & 35.90 \\
			& SPO+ (cached) & 159{,}213.45 & 98.10 \\
			& SPO+ (non-cached) & 159{,}032.17 & 1{,}102.86 \\
			& $\restro{SPO+}$ (exact $\eta$) & 157{,}237.15 & 106.83 \\
			& $\restro{SPO+}$ (quantile $\eta$) & 157{,}176.50 & 97.88 \\
			\midrule
			\multirow{4}*{Heterogeneous}
			& MSE & 242{,}105.05 & 35.41 \\
			& SPO+ (cached) & 238{,}957.60 & 103.35 \\
			& $\restro{SPO+}$ (exact $\eta$) & 231{,}704.70 & 108.97 \\
			& $\restro{SPO+}$ (quantile $\eta$) & 227{,}848.71 & 107.40 \\
			\bottomrule
	\end{tabular}}
	{Mean over 5 random seeds. Non-cached SPO+ was run only on the original data.}
\end{table}
\section{Conclusion}\label{sec:conclusion}

We proposed a risk-sensitive DFL framework, RESTRO, that turns existing DFL methods such as SPO+ into variants that minimize the CVaR of regret. The framework integrates with standard DFL pipelines by reweighting gradients at the minibatch level, without changes to the underlying solver or model architecture. A key advantage is that it relies on point predictions of the uncertain parameters followed by a \emph{deterministic} optimization problem. This distinguishes it from prior risk-sensitive approaches that solve more complex stochastic or robust optimization problems to account for tail risk. For the SPO+ instantiation, we showed that the RESTRO objective controls an upper bound on the population CVaR of the true regret. Empirically, RESTRO substantially reduces tail regret on knapsack, portfolio, shortest-path, and energy-aware scheduling problems, especially when the cost parameters exhibit heterogeneous tail behavior, and it remains competitive when they do not.

\paragraph{Limitations and future work.}
Although the reweighting mechanism can be combined with any DFL surrogate, its benefit depends on the underlying loss: the improvements are strongest and most consistent for SPO+, which is also the instantiation covered by our theoretical guarantee. Extending the guarantee to surrogates without a pointwise regret upper bound, such as IMLE and PFY, and to methods that differentiate through smoothed optimization problems is an interesting direction. More broadly, this paper considers uncertainty in the objective parameters of CO problems. Uncertainty in the constraint parameters is more challenging, because it affects both feasibility and optimality, and recent DFL works have begun to study this setting~\citep{hu2023predict+, hu2023twostage, mandi2026feasibilityaware}. In future work, we aim to extend RESTRO to uncertain constraint parameters.


\bibliographystyle{informs2014}
\bibliography{myref}

\end{document}